\documentclass[10pt,twocolumn,letterpaper]{article}

\usepackage{cvpr}              % To produce the CAMERA-READY version
\usepackage[accsupp]{axessibility}
\usepackage{graphicx}
\usepackage{amsmath}
\usepackage{amssymb}
\usepackage{booktabs}
\usepackage{multirow}
\usepackage[pagebackref,breaklinks,colorlinks,bookmarks=false]{hyperref}
\usepackage{bm}
\usepackage{bbm}
\usepackage{makecell} 
\usepackage{booktabs}  
\usepackage{color}

\usepackage[capitalize]{cleveref}
\crefname{section}{Sec.}{Secs.}
\Crefname{section}{Section}{Sections}
\Crefname{table}{Table}{Tables}
\crefname{table}{Tab.}{Tabs.}

\def\cvprPaperID{2} % *** Enter the CVPR Paper ID here
\def\confName{CVPR}
\def\confYear{2023}

\begin{document}

\title{Exposing Fine-Grained Adversarial Vulnerability of Face Anti-Spoofing Models}

\author{Songlin Yang\textsuperscript{1,2}, Wei Wang\textsuperscript{2,\thanks{Corresponding author.} }, Chenye Xu\textsuperscript{3}, Ziwen He\textsuperscript{1,2}, Bo Peng\textsuperscript{2}, Jing Dong\textsuperscript{2}\\
\textsuperscript{1}School of Artificial Intelligence, University of Chinese Academy of Sciences\\
\textsuperscript{2}Center for Research on Intelligent Perception and Computing, CASIA\quad\textsuperscript{3}SenseTime Research\\
{\tt\small {yangsonglin2021@ia.ac.cn, xuchenye@sensetime.com, $\{$wwang, bo.peng, jdong$\}$@nlpr.ia.ac.cn}}}

% \author{Songlin Yang, Wei Wang, Chenye Xu, Ziwen He, Bo Peng, and Jing Dong\\
% \and
\maketitle

%%%%%%%%% ABSTRACT
\begin{abstract}
   Face anti-spoofing aims to discriminate the spoofing face images (e.g., printed photos and replayed videos) from live ones. However, adversarial examples greatly challenge its credibility, where adding some perturbation noise can easily change the output of the target model. Previous works conducted adversarial attack methods to evaluate the face anti-spoofing performance without any fine-grained analysis that which model architecture or auxiliary feature is vulnerable. To handle this problem, we propose a novel framework to expose the fine-grained adversarial vulnerability of the face anti-spoofing models, which consists of a multitask module and a semantic feature augmentation (SFA) module. The multitask module can obtain different semantic features for further fine-grained evaluation, but only attacking these semantic features fails to reflect the vulnerability which is related to the discrimination between spoofing and live images. We then design the SFA module to introduce the data distribution prior for more discrimination-related gradient directions for generating adversarial examples. And the discrimination-related improvement is quantitatively reflected by the increase of attack success rate, where comprehensive experiments show that SFA module increases the attack success rate by nearly 40$\%$ on average. We conduct fine-grained adversarial analysis on different annotations, geometric maps, and backbone networks (e.g., Resnet network). These fine-grained adversarial examples can be used for selecting robust backbone networks and auxiliary features. They also can be used for adversarial training, which makes it practical to further improve the accuracy and robustness of the face anti-spoofing models. Code: \small\url{https://github.com/Songlin1998/SpoofGAN}
\end{abstract}

\section{Introduction}

Face anti-spoofing~\cite{yu2022deep,wu2021dual} is significantly important to the credibility of face recognition systems, which aims to determine whether a presented face is live or spoofing. If the face anti-spoofing part is unreliable, malicious attackers can use photos and videos of your face to unlock your mobile phone or other biometric authentication systems. The researchers~\cite{abdullakutty2021review} adopted different backbone networks (e.g., Resnet~\cite{he2016deep} and Transformer~\cite{liu2021swin,wang2022face}) and auxiliary information~\cite{zhang2020celeba} (e.g., depth maps and facial attributes), and proposed highly complicated face anti-spoofing models~\cite{yu2022deep}. These models improved the performance of face anti-spoofing to high accuracy. However, the emergence of adversarial examples~\cite{huang2017adversarial,akhtar2018threat,kong2021survey} poses a fatal threat to face anti-spoofing models, which can easily mislead the target model, making it output a wrong classification result with high confidence. For example, an image of your photo could have been classified as spoofing input, but this image will be classified as live input, after the modification of the adversarial attacks.

Previous works~\cite{zhang2020adversarial,wu2020defense,yang2021sparse,mao2021research} have tried to conduct adversarial attacks to these models, which only revealed the adversarial vulnerability of these face anti-spoofing models. But these attacks~\cite{carlini2019evaluating,jiang2019avoiding,dong2020benchmarking,rauber2020foolbox,eger2020hero,tong2021facesec,liu2022practical} entangled discrimination features (i.e, spoofing-live classification) with auxiliary features and failed to expose the fine-grained adversarial analysis. In other words, they are not able to figure out which part of the target face anti-spoofing model is vulnerable, especially the models using several auxiliary features to assist in discrimination. This makes it difficult for researchers to improve the adversarial robustness of face anti-spoofing models while maintaining the accuracy performance simultaneously.
    
To expose the fine-grained adversarial vulnerability of face anti-spoofing models, we propose a novel framework that consists of a multitask module and a semantic feature augmentation (SFA) module. The multitask module is a backbone network with several branches to obtain different semantic features corresponding to different auxiliary information for spoofing-live discrimination of face anti-spoofing models. This architecture is flexible to evaluate different auxiliary information and backbone networks systematically. However, only attacking these semantic features without spoofing-live prior fails to reflect the spoofing-live discrimination correlation. This means the low attack success rate towards the spoofing-live classification via attacking
the semantic features. We then design the SFA module to introduce the data distribution prior for more discrimination-related gradient directions. The discrimination-related improvement is quantitatively reflected by the increase of attack success rate and our SFA module boosts the attack success rate of adversarial examples even for the one-step attack such as FGSM~\cite{huang2017adversarial}, which provides an efficient way to generate a large number of adversarial examples for quantitative experiments. This fine-grained vulnerability can be used for selecting robust backbone networks and auxiliary features, which provides a significantly important tool for model optimization, further improving the accuracy and robustness of the face anti-spoofing models.
    
   \textbf{The main contributions of this work are as follows:}
     \begin{itemize}
        \item We propose a novel framework to systematically expose the fine-grained adversarial vulnerability of face anti-spoofing models.
        \item We propose a semantic feature augmentation  (\textbf{SFA}) module to obtain discrimination-related gradient directions, increasing the attack success rate by nearly 40\% on average.
        \item We conduct comprehensive experiments from three perspectives, which are annotations (facial attributes, spoofing types, and illumination), geometric information (depth and reflection maps), and backbone networks.
    
    \end{itemize}

    \begin{figure*}
        \centering
        \includegraphics[scale=0.59]{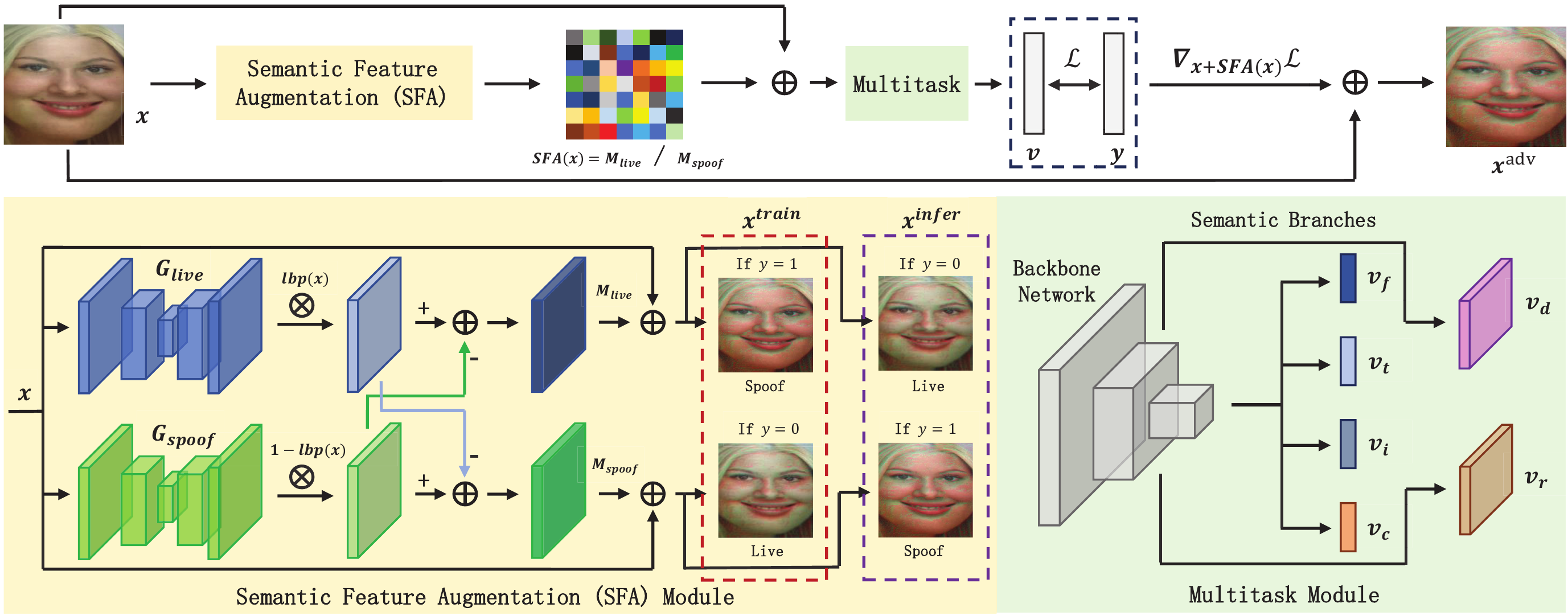}
        \caption{The framework for exposing the fine-grained adversarial vulnerability of face anti-spoofing models. The semantic feature augmentation (SFA) module aims to obtain the live/spoofing activation map $\bm{M}_{live}$/$\bm{M}_{spoof}$, and add the map to the input image. The multitask module is adopted to obtain the semantic features of the input image. With the SFA and multitask module, we are able to conduct adversarial attacks towards different backbone networks and auxiliary information for fine-grained adversarial analysis in the context of face anti-spoofing.}
        \label{framework}
    \end{figure*}

\section{Related Work}
\subsection{Face Anti-Spoofing}
    Face anti-spoofing is an important guarantee for the reliability of face recognition systems, especially in some security scenarios~\cite{yu2022deep,wu2021dual,yu2022deep}. In early research, face anti-spoofing algorithms were based on handcrafted features, such as LBP~\cite{yang2013face}, HoG~\cite{schwartz2011face}, and SURF~\cite{boulkenafet2016face}. Temporal features like eye-blinking~\cite{pan2007eyeblink} and lip motion~\cite{kollreider2007real} also received attention. Methods based on different color spaces have also been proposed, such as HSV~\cite{boulkenafet2016face}, YCbCr~\cite{boulkenafet2016face} and Fourier spectrum~\cite{li2004live}. With the popularity of methods based on deep learning, CNNs have been used for feature extraction and classification, and these CNN-based methods achieved excellent performance~\cite{yu2022deep,belli2022personalized,liu2022adversarial}, nearly 100\% accuracy in academic datasets. Auxiliary information including annotations~\cite{zhang2020celeba} (e.g., facial attributes, spoofing types, and illumination) and geometric maps~\cite{kim2019basn} (depth and reflection maps), were studied to assist the binary classification. However, the emergence of adversarial attacks~\cite{zhang2020adversarial,wu2020defense,yang2021sparse,mao2021research} exposed the vulnerability of face anti-spoofing. Different auxiliary information and backbone networks can indeed improve the classification accuracy of live and spoofing, but for this task related to security, its adversarial robustness should be studied.
    
\subsection{Adversarial Attack}
    The models based on deep learning are vulnerable to adversarial attacks, which is a popular research concern in recent years~\cite{akhtar2018threat,zhang2022towards}. By adding the imperceptible noise to the original data, adversarial examples can mislead the classification easily~\cite{dong2020benchmarking}. Adversarial attacks can be divided into white-box and black-box attacks. White-box attacks~\cite{huang2017adversarial,carlini2017towards,madry2017towards} generated the adversarial perturbation by obtaining the gradient of the model, while black-box attacks~\cite{ma2021simulating,cina2022black} focus on the transferability~\cite{xie2019improving} of adversarial examples (i.e., using the adversarial examples generated on one model can be used to attack other models).

    Besides being studied as a problem~\cite{rice2020overfitting,pang2020bag,yang2021systematical}, adversarial examples can also be used as a tool to expose the vulnerability of the model, as well as measuring the impact of data distribution and network architecture on the adversarial robustness~\cite{tong2021facesec,jiang2019avoiding,carlini2019evaluating,dong2020benchmarking,rauber2020foolbox,eger2020hero}. However, previous methods~\cite{zhang2020adversarial,wu2020defense} towards the adversarial attacks of face anti-spoofing merely tackle the fine-grained adversarial vulnerability from the perspective auxiliary information.

    \noindent
    \textbf{Physical Adversary vs. Face Spoofing Attack} Some literature~\cite{xiao2021improving,komkov2021advhat,cheng2022physical} has explored the physical adversarial attacks in real-world scenes, making adversarial examples generated in the digital domain can still mislead the target model after being printed. However, the aim of face anti-spoofing models is to distinguish spoofing images captured from the physical domain and real-world diversity has been considered in the dataset, which means the face spoofing attack itself is a physical attack. Our goal is to quantitatively evaluate the template discriminant features of face anti-spoofing models (i.e., efficiently test whether face anti-spoofing models can robustly classify a large number of input images in the digital domain). So we focus on digital adversarial attacks first.
    
\subsection{Class Activation Map (CAM)}
    Explicability of classification decisions and localization made by the neural networks are obtained via the computation of class activation map (CAM)~\cite{selvaraju2017grad}. Furthermore, CAM and its following works~\cite{chattopadhay2018grad,wang2020score,ramaswamy2020ablation,muhammad2020eigen,jiang2021layercam} visualize class-relevant features in the form of a heatmap, which is the reason why we select CAM~\cite{selvaraju2017grad} as a visualization tool.
    
\section{Method}
    We present our framework for exposing the fine-grained adversarial vulnerability of face anti-spoofing models in Fig.~\ref{framework}. We first adopt the semantic feature augmentation (SFA) module (Sec.~\ref{3.2}) to obtain the live/spoofing activation map and add the map to the input image. And then, we feed the processed image into the multitask module to obtain the semantic features. Finally, we conduct fine-grained adversarial attacks (Sec.~\ref{3.3}) toward different backbone networks and auxiliary information. Before introducing the details of our proposed method, we first take a one-step adversarial attack, FGSM \cite{huang2017adversarial}, as a representative method to introduce the generation process of adversarial examples in Sec.~\ref{3.1}.

\subsection{Preliminary}
\label{3.1}
    The adversarial example $\bm{x}^{adv}$ is the image modified by the adversarial noise, and it can make the target model predict $f(\bm{x}^{adv})=\bm{v}$ while such output is different from its label $\bm{y}$. FGSM~\cite{huang2017adversarial} is a one-step attack, which is formally defined by
    \begin{equation}
        \bm{x}^{adv} = \bm{x} + \epsilon\cdot sign(\nabla_{\bm{x}}\; \mathcal{L}(\bm{v},\bm{y})),
    \end{equation}
    where $\bm{x}$, $\epsilon$, and $\mathcal{L}(\cdot)$ denote the input image, max perturbation scale, and loss function. The $sign(\cdot)$ is a mathematical function that extracts the sign of a real number or vector.

\subsection{Semantic Feature Augmentation (SFA)}
\label{3.2}
    We design our SFA module from data and model perspectives: Previous works augmented data without class information or with just their corresponding label~\cite{mohamed2021face,chen2022learning}, while SFA considers both contrastive labels. Taking a live image $\bm{x}\in \mathbbm{R}^{3 \times H\times W}$ as an example, we augment the live features of a live image, by adding live activation map $\bm{M}_{live}\in \mathbbm{R}^{1 \times H\times W}$ to strengthen live features and subtracting spoofing activation map $\bm{M}_{spoof}\in \mathbbm{R}^{1 \times H\times W}$ to weaken spoofing features, where $H$ and $W$ are the height and width of the input image. Moreover, CNN-based models are biased to texture changes~\cite{geirhos2018imagenet}, so we adopt $LBP$~\cite{chingovska2012effectiveness} to manipulate textures to further increase the attack success rate. These are reasons why SFA can make processed images more semantic-aware to live/spoofing features and achieve better results than class activation map~\cite{selvaraju2017grad} and vanilla data augmentation methods (e.g., flip and rotation). Next, we introduce each part of the pipeline of SFA module, as shown in Fig.~\ref{framework}:
    
    \noindent
   \textbf{Generator.} Two variational autoencoders (VAE)~\cite{kingma2013auto}, denoted as $G_{live}$ and $G_{spoof}$, are used for generating live and spoofing activation maps ($\bm{M}_{live}$ and $\bm{M}_{spoof}$) respectively. Then, the activation maps output by $G_{live}$ and $G_{spoof}$ will be multiplied by $LBP$~\cite{chingovska2012effectiveness} through Hadamard product denoted as $lbp(\cdot)$ and $1-lbp(\cdot)$. Thus, region-wise manipulation of texture can be achieved. To be specific, for the input image $\bm{x}$, the activation maps $\bm{M}_{live}$ and $\bm{M}_{spoof}$ can be obtained by the following formula:
   
    \begin{equation}
      \bm{M}_{live}(\bm{x}) = lbp(\bm{x})\odot G_{live}(\bm{x})-[1-lbp(\bm{x})]\odot G_{spoof}(\bm{x}),
    \end{equation}
    \begin{equation}
      \bm{M}_{spoof}(\bm{x}) = -lbp(\bm{x})\odot G_{live}(\bm{x})+[1-lbp(\bm{x})]\odot G_{spoof}(\bm{x}),
    \end{equation}
    
    \noindent
    \textbf{Discriminator.} The pretrained face anti-spoofing model is chosen to be the discriminator, denoted as $D$. The parameters of this target model are fixed in the process of optimizing the generator.
    
    \noindent
    \textbf{Training Strategy and Loss Function.} The modified image $\bm{x}^{train}_{i}$ for optimizing $G_{live}$ and $G_{spoof}$, can be generated by Eq.~\ref{training}. Note that input image $\bm{x}_{i}$ is added by activation map opposite to its label:
    \begin{equation}
       \bm{x}^{train}_{i} = \bm{x}_{i} + y_{i}\cdot \bm{M}_{live}(\bm{x}_{i})+(1-y_{i})\cdot \bm{M}_{spoof}(\bm{x}_{i}),
       \label{training}
    \end{equation}
    where $[\bm{x}_{i},y_{i}]_{i=1}^{N}$ is a batch of training data with labels. Note that $y_{i}$=0 if the input image is live, otherwise $y_{i}$=1. 
    
    The training objective is to make the $\bm{x}^{train}_{i}$, inferred by the discriminator $D$, consistent with the opposite label. As shown below, the Binary Cross Entropy is adopted:
    \begin{equation}
       \mathcal{L}_{1} = \frac{-1}{N}\sum_{i=1}^{N}[(1-y_{i})\cdot log(D(\bm{x}^{train}_{i}))+y_{i}\cdot log(1-D(\bm{x}^{train}_{i}))],
    \end{equation}
    
    To better introduce our method and experimental results in the following sections, we denote the Semantic Feature Augmentation module as $SFA(\cdot)$. Note that $SFA(\bm{x})$ generate the activation map same as the label of $x$:
    \begin{equation}
       SFA(\bm{x}) = y\cdot \bm{M}_{spoof}(\bm{x})+(1-y)\cdot \bm{M}_{live}(\bm{x}),
      \label{sfa}
    \end{equation}
    \begin{equation}
       \bm{x}^{infer} = \bm{x} + SFA(\bm{x}).
       \label{inference}
    \end{equation}

    \begin{table}
     \footnotesize
        \centering
        \caption{The attack success rates of different adversarial attacks without and with SFA.}
        \begin{tabular}{ccc}
                \Xhline{1pt}
                \multirow{2}{*}{Adversarial Attack}&\multicolumn{2}{c}{Attack Success Rate$\;\uparrow$} \\
                \cline{2-3}
                \multirow{2}{*}{~}& without SFA & with SFA\\
                \Xhline{0.5pt}
                None & 0.0002 & \textbf{0.7129}\\
                FGSM~\protect\cite{huang2017adversarial}  ($\epsilon=0.06$) & 0.6578 & \textbf{0.9046}\\
                FGSM~\protect\cite{huang2017adversarial}  ($\epsilon=0.1$) & 0.6899 & \textbf{0.9120}\\
                FGSM~\protect\cite{huang2017adversarial}  ($\epsilon=0.2$) & 0.6983 & \textbf{0.9161}\\
                C\&W~\protect\cite{carlini2017towards}  ($\epsilon=0.06$) & 0.6925 & \textbf{0.8812}\\
                 C\&W~\protect\cite{carlini2017towards}  ($\epsilon=0.1$) & 0.7239 & \textbf{0.9122}\\
                 C\&W~\protect\cite{carlini2017towards}  ($\epsilon=0.2$) & 0.8039 & \textbf{0.9258}\\
               Spatial~\protect\cite{xiao2018spatially}  ($\epsilon=0.06$) & 0.4786 & \textbf{0.9443}\\
                PGD~\protect\cite{madry2017towards} ($\epsilon=0.1$) & 0.7045 & \textbf{0.9471} \\
                Sparse~\protect\cite{croce2019sparse}  ($\epsilon=0.5$) & 0.1261 & \textbf{0.7832}\\
                \Xhline{1pt}
            \end{tabular}
        \label{t1}
    \end{table}
    
\subsection{Fine-Grained Adversarial Attack}
\label{3.3}

    Many previous works~\cite{zhang2020celeba,kim2019basn} used auxiliary information (e.g. illumination maps) to help face anti-spoofing models learn the discriminant boundary. We perturb the specific auxiliary feature and evaluate the effect of such perturbation, to figure out which part really affects the discriminant boundary of spoofing/live. The multitask network aims to get learned auxiliary features for generating fine-grained adversarial examples. Such a 'Backbone + Branches' structure is suitable for studying the relationships between auxiliary information, backbone networks, and spoofing/live discriminant features.

    \begin{figure}
       \centering
       \includegraphics[scale=0.41]{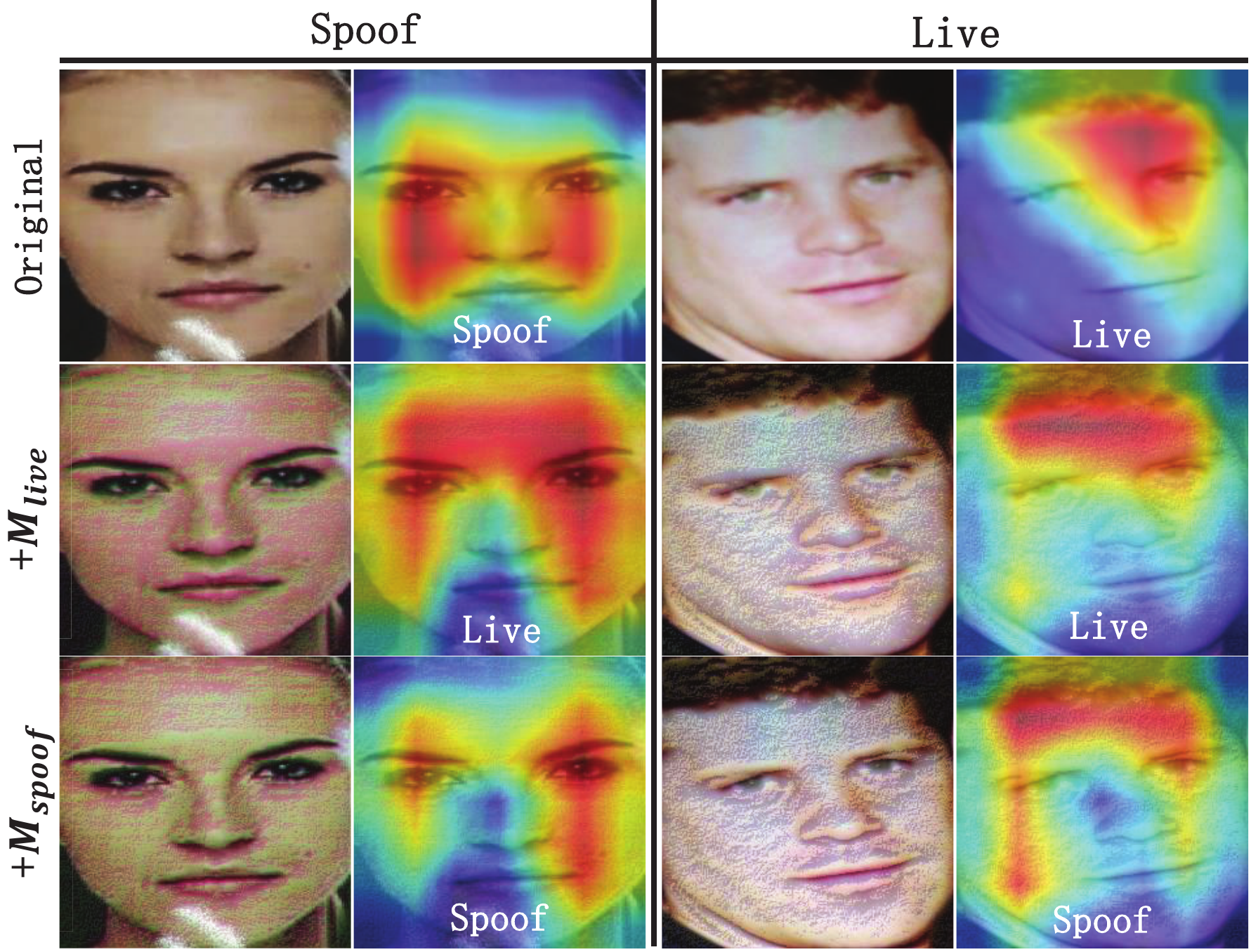}
       \caption{Visualization and perturbation effect on the classification of live and spoofing maps generated by SFA. An original spoofing (replay attack) and live sample with their class activation maps (CAM) are shown in 1st and 2nd columns, while 2nd and 3rd rows are the results added by $\bm{M}_{live}$ and $\bm{M}_{spoof}$ respectively. The spoofing activation region tends to distribute around the face in the form of vertical lines, while the live activation region distributes in the forehead area of the face.}
       \label{1}
    \end{figure}

    In order to fully mine the impact of data annotation and auxiliary information on the adversarial robustness of face anti-spoofing, we adopt the multitask network with auxiliary information~\cite{zhang2020celeba} as a basic model. As shown in Fig.~\ref{framework}, the backbone network is marked as gray, and the last logits are output to four fully-connected layers, to obtain the vector $\bm{v}_{f}$, $\bm{v}_{t}$, $\bm{v}_{i}$ and $\bm{v}_{c}$, which represents facial attribute, spoofing type, illumination and binary classification of live and spoofing. The three annotation vectors are compared with ground-truth label $\bm{y}_{f}$,  $\bm{y}_{t}$ and $\bm{y}_{i}$, thus we get the semantic loss function $\mathcal{L}_{a}$ as follows:
    \begin{equation}
        \mathcal{L}_{a} = \lambda_{f}\cdot \mathcal{L}_{f}(\bm{v}_{f},\bm{y}_{f})+\lambda_{t}\cdot \mathcal{L}_{t}(\bm{v}_{t},\bm{y}_{t})+\lambda_{i}\cdot \mathcal{L}_{i}(\bm{v}_{i},\bm{y}_{i}),
    \end{equation}
    where $\mathcal{L}_{t}$ and $\mathcal{L}_{i}$ use Softmax Cross Entropy loss, and $\mathcal{L}_{f}$ uses Binary Cross Entropy loss. 
    
    Furthermore, reflection map $\bm{v}_{r}$ and depth map $\bm{v}_{d}$ are captured by two geometric map generators. According to \cite{zhang2020celeba}, for the depth map of the sample labeled as live, its ground truth $\bm{y}_{d}$ is obtained by PRNet~\cite{feng2018joint}, while the depth map of the sample labeled as spoofing is zero. The method proposed in \cite{zhang2018single} is adopted to generate the ground truth of the reflection map of the sample labeled as spoofing, and the reflection map of the sample labeled as live is zero. Therefore, we can get the following geometric loss function $\mathcal{L}_{g}$:
    \begin{equation}
        \mathcal{L}_{g} =\lambda_{d}\cdot \mathcal{L}_{d}(\bm{v}_{d}, \bm{y}_{d})+\lambda_{r}\cdot \mathcal{L}_{r}(\bm{v}_{r},\bm{y}_{r}),
    \end{equation}
    where $\mathcal{L}_{d}$ and $\mathcal{L}_{r}$ are Mean Square Error loss. We use Softmax Cross Entropy loss as $\mathcal{L}_{c}$, thus we can get the final optimization objective:
    \begin{equation}
        \mathcal{L}_{2} = \mathcal{L}_{c}(\bm{v}_{c},\bm{y}_{c}) + \mathcal{L}_{a} + \mathcal{L}_{g},
        \label{loss2}
    \end{equation}
    
    Then, from the perspective of auxiliary information (annotation vectors and geometric maps), we attack different parts of the last layer and analyze the adversarial robustness of the target model. The adversarial vulnerability of different auxiliary information can be illustrated by the change in classification accuracy. However, only attacking these semantic features fails to reflect the discrimination-related vulnerability, which means a low attack success rate towards the spoofing-live classification via attacking the semantic features. So adversarial perturbation is generated based on the images modified by SFA to obtain gradient directions related to spoofing-live discrimination.
    
    We adopt FGSM\cite{huang2017adversarial} as the basic method to present the formulation of generating these fine-grained adversarial examples:
    \begin{equation}
        \bm{x}^{adv}_{s} = \bm{x} + \epsilon\cdot sign(\nabla_{\bm{x}+SFA(\bm{x})}\; \mathcal{L}_{s}(\bm{v}_{s},y_{s})),
    \end{equation}
    where $s\in\{f, t, i, d, r, c\}$ respectively represent the facial attribute, spoofing type, illumination, depth map, reflection map, and classification. $\epsilon$ indicates the step of the attack. 
    
    Furthermore, replacing the backbone network with different state-of-art models can provide us with a new perspective to study the adversarial vulnerability of face anti-spoofing models, from the effects of backbone structures.

    \begin{table}
    \footnotesize
        \centering
        \caption{The attack success rates (ASR) of FGSM~\protect\cite{huang2017adversarial} with SFA, data augmentation and class activation map. Note that `Ensemble' means ensemble with different data augmentation operations.}
        \begin{tabular}{ccccc}
        \Xhline{1pt}
        \multicolumn{2}{c}{Data Augmentation}& &\multicolumn{2}{c}{Class Activation Map}\\
        \cline{1-2} \cline{4-5}
        Method & ASR$\;\uparrow$ &&  Method & ASR$\;\uparrow$ \\
        \Xhline{0.5pt}
        None & 0.6578 & & None & 0.6578\\
        Vertical Flip & 0.7313&  & Gradcam~\protect\cite{selvaraju2017grad} & 0.7375\\
        Horizontal Flip & 0.7063&  & Gradcam++~\protect\cite{chattopadhay2018grad} & 0.7625\\
        Rotation(30°) & 0.7563&  & Scorecam~\protect\cite{wang2020score} & 0.2875\\
        Brightness & 0.6938&  & Ablationcam~\protect\cite{ramaswamy2020ablation} & 0.7375\\
        Hue & 0.7250&  & Eigencam~\protect\cite{muhammad2020eigen} & 0.8000\\
        Ensemble & 0.8152&  & Layercam~\protect\cite{jiang2021layercam} & 0.7625\\
        
        SFA & \textbf{0.9046}&  & SFA & \textbf{0.9046}\\
        \Xhline{1pt}
        \end{tabular}
        \label{t2}
    \end{table}
    \begin{figure}
       \centering
       \includegraphics[scale=0.43]{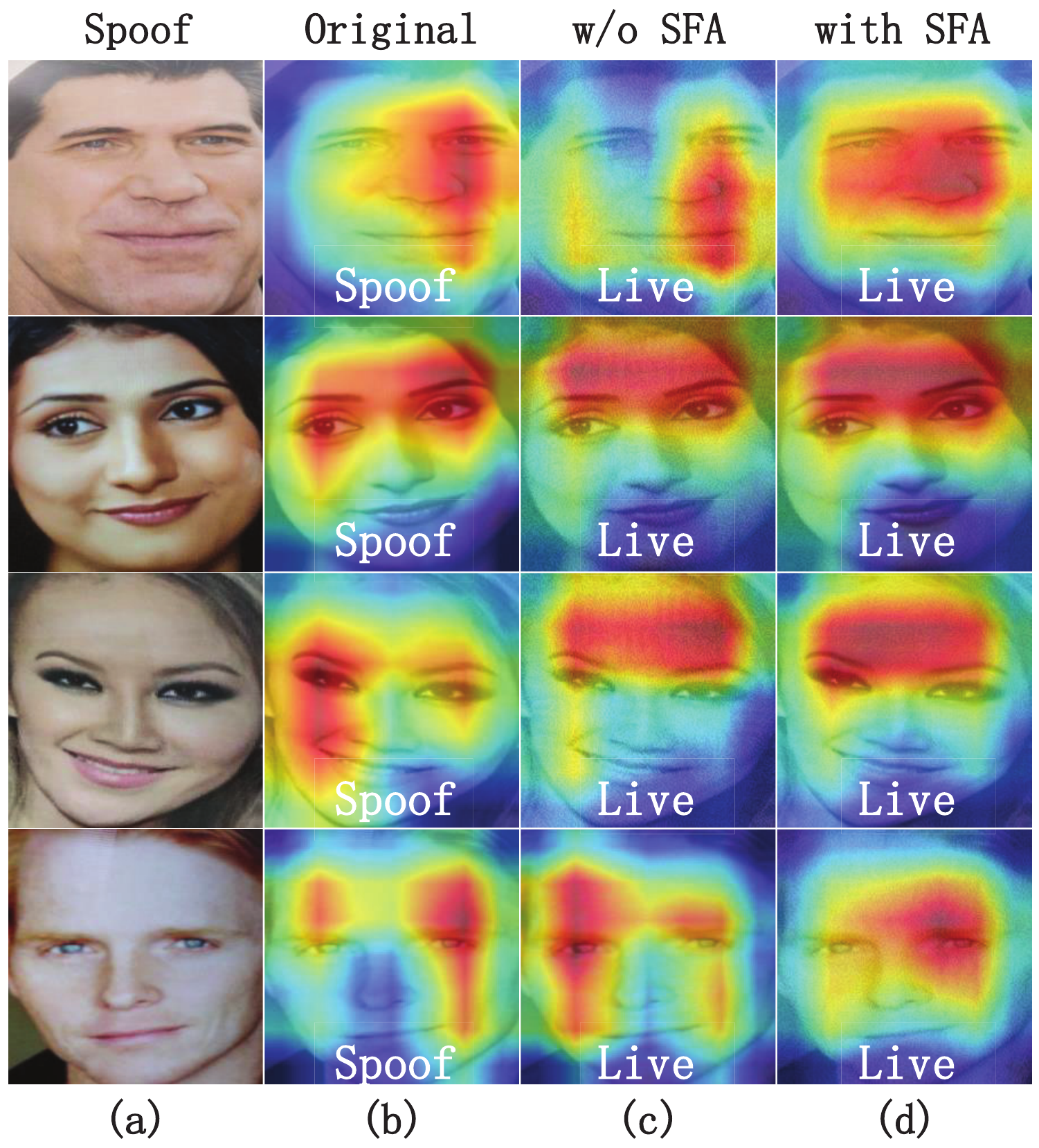}
       \caption{ Differences of adversarial attacks without and with SFA module. (a) Original spoofing images obtained from CelebA-spoof~\protect\cite{zhang2020celeba}. (b) CAM of the original images. (c) CAM of the adversarial images without SFA. (d) CAM of the adversarial images with SFA (completely different from Fig. 4(b)). Note that (c) and (b) are perturbed by the FGSM~\protect\cite{huang2017adversarial} with the same attack step ($\epsilon=0.1$) and backbone network (Resnet-18~\cite{he2016deep}). SFA makes the adversarial perturbation more concentrated on the semantic features and boosts adversarial attacks, so attacks with SFA do change discrimination-related information of input images. Such effects and differences are illuminated by CAM, which can explicitly locate the classification decision.}
     
       \label{4}
    \end{figure}
\section{Experiments}
\subsection{Experimental Settings}
\noindent
\textbf{Network Architectures.} We adopt the VAE architecture~\cite{kingma2013auto} as $G_{live}$ and $G_{spoof}$ of SFA module, which has five hidden units with $\{32, 64, 128, 256, 512\}$ dimensions. Each hidden unit consists of vanilla 2D convolution, batch normalization, and LeakyReLU activation layer. In multitask module, the two geometric map generators consist of a Conv $3\times3$ followed by an upsample to $14\times14$. We set $\lambda_{f}=1$, $\lambda_{t}=0.1$, $\lambda_{i} = 0.01$, $\lambda_{d}=0.1$ and $\lambda_{r}=0.1$.

\noindent
\textbf{Dataset.} The purpose of our paper is to expose the adversarial vulnerability of face anti-spoofing from the fine-grained perspective and figure out which part of the target models is vulnerable. So our experiments use CelebA-spoof~\cite{zhang2020celeba} as the dataset, which has three significant advantages: (a) Large-Scale: CelebA-spoof comprises 625,537 pictures of 10,177 subjects. (b) Diversity: The spoofing images are captured from 8 scenes (2 environments $\times$ 4 illumination conditions) with more than 10 sensors. (c) Annotation richness: CelebA-spoof contains 10 spoofing type annotations, as well as the 40 attribute annotations inherited from the original CelebA~\cite{liu2018large} dataset. These three advantages are helpful for us to construct the required experimental scenarios, and the experimental results based on this large-scale dataset are more general. We randomly sample 600,000 images from different categories to train the SFA module, and we randomly sample 10,000 images from the left data in every evaluation.

\noindent
\textbf{Metrics.} The attack success rate (ASR) and classification accuracy are metrics for the performance of adversarial attacks. The results predicted by the target model are correct when the output after Softmax is the same as the ground truth. Mean Square Error is taken to measure the change of each feature output. Each quantitative result is tested at least three times and then averaged.
   
\subsection{Effectiveness Analysis of SFA Module}
    
    The semantic feature augmentation (SFA) module is designed to provide better gradients related to the spoofing-live discrimination. Its characteristic is to learn the activation maps of the target model towards both classes of live and spoofing data, by constructing positive and negative samples weighted by the texture filter. With the help of SFA, the adversarial perturbation can be added to more semantic-aware to live and spoofing features. In this way, it can make gradients more related to the spoofing-live decision boundary, thus improving the attack success rate of the fine-grained adversarial attacks.

     \begin{table}
     \footnotesize
        \centering
        \caption{The ablation study of SFA without and with LBP using different adversarial attacks.}
        
        \begin{tabular}{ccc}
            \Xhline{1pt}
            \multirow{2}{*}{Attack Methods} &\multicolumn{2}{c}{Attack Success Rate$\;\uparrow$}\\
            \cline{2-3}
            ~& without LBP & with LBP \\
            \Xhline{0.5pt}
            FGSM~\protect\cite{huang2017adversarial}  ($\epsilon=0.1$) & 0.7121 & \textbf{0.9120$\;$}\\
            C\&W~\protect\cite{carlini2017towards} 
 ($\epsilon=0.1$) & 0.8403 & \textbf{0.9122$\;$}\\
            Spatial~\protect\cite{xiao2018spatially}  ($\epsilon=0.06$) & 0.8001 & \textbf{0.9443$\;$}\\
            PGD~\protect\cite{madry2017towards} 
 ($\epsilon=0.1$) & 0.8832 & \textbf{0.9471$\;$} \\
            Sparse~\protect\cite{croce2019sparse} 
 ($\epsilon=0.5$) & 0.6755 & \textbf{0.7832$\;$}\\
            \Xhline{1pt}
        \end{tabular}
        \label{t3}
    \end{table}

    To thoroughly evaluate the effectiveness of SFA, in this section, we first show that SFA can generate two different maps for one image (See Fig.~\ref{1}), and study the effect of SFA on different adversarial attacks, to verify that SFA can improve the attack success rate (See Fig.~\ref{4} and Tab.~\ref{t1}). Then, we compare SFA with different methods of class activation map (CAM) and data augmentation (See Tab.~\ref{t2}), to illustrate the advantages of SFA over previous methods. Finally, ablation experiments on the texture filter are carried out to show its necessity (See Tab.~\ref{t3}).
    
    \noindent
    \textbf{Contrastive Activation Maps of Live and Spoofing.} SFA can generate two different maps for one image, corresponding to live and spoofing respectively, as shown in Fig.~\ref{1}. According to numerous empirical observations on the visualization of experimental results, spoofing activation tends to distribute at the edge and has the law of a quadrilateral grid. For live activation, the inner region of the face is stronger, especially the forehead. Such contrastive activation maps extract the discrimination information, which can be adopted for data augmentation to boost the adversarial attacks. We use the $SFA(\cdot)$ denoted in Eq.~\ref{sfa} to add an activation map with the same label to the input image to obtain better gradient directions to generate adversarial examples.

     \begin{table}
     \footnotesize
        \centering
        \caption{The accuracy of Resnet-18~\protect\cite{he2016deep} when different annotation vectors and geometric maps are attacked.}
        \begin{tabular}{ccc}
            \Xhline{1pt}
            \multirow{2}{*}{Attacked Parts} &\multicolumn{2}{c}{Attack Success Rate$\;\uparrow$}\\
            \cline{2-3}
            ~& without SFA & with SFA \\
            \Xhline{0.5pt}
            Facial Attribute & 0.6341 & 0.6201\\
            Spoofing Type & 0.6455 & \textbf{0.7559}\\
            Illumination & 0.6475 & \textbf{0.7679}\\
            
            Depth Map & 0.0041 & \textbf{0.6694}\\
            Reflection Map & 0.0001 & \textbf{0.6657}\\
            
            Classification & 0.6578 & \textbf{0.9046} \\
            \Xhline{1pt}
        \end{tabular}
        \label{t4}
    \end{table}

 \begin{figure}
        \centering
        \includegraphics[scale=0.6]{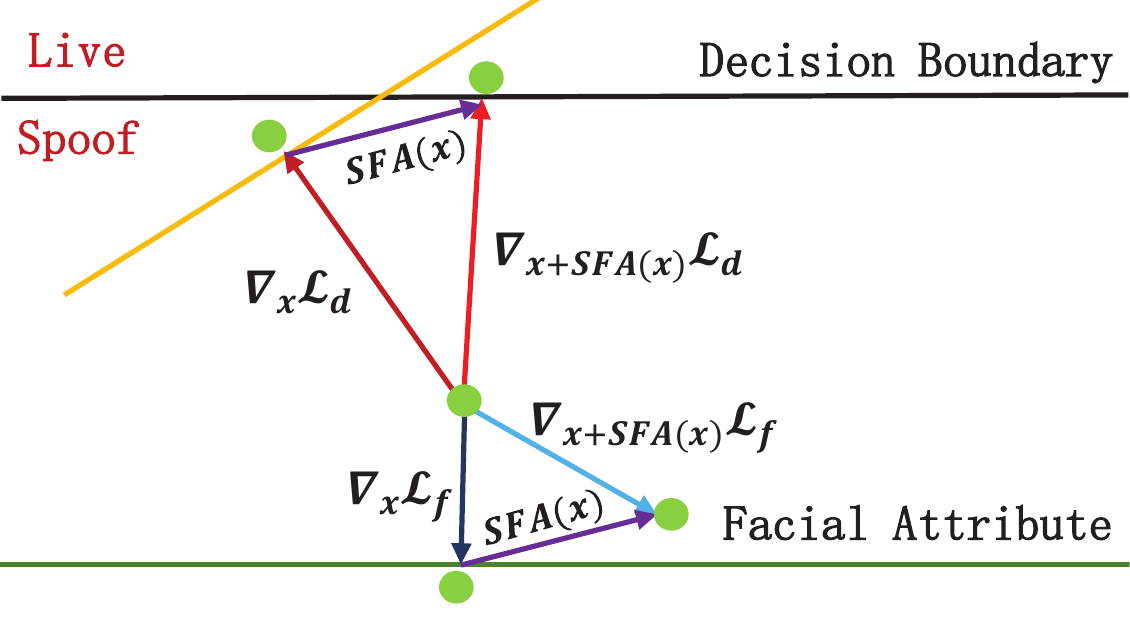}
        \caption{The interpretation of differences between facial attribute boundary and depth map boundary.}
        \label{6}
    \end{figure}
    
    \noindent
    \textbf{Different Adversarial Attacks with SFA.} As shown in Tab.~\ref{t1}, the attack success rates of different adversarial attacks~\cite{huang2017adversarial,madry2017towards,croce2019sparse,carlini2017towards,xiao2018spatially} based on Resnet-18~\cite{he2016deep} have improved a lot with SFA. To qualitatively demonstrate the influence of the SFA module on discrimination of the target face anti-spoofing model, as shown in Fig.~\ref{4}, we then visualize a spoofing example with its adversarial examples generated without and with SFA based on the Resnet-18~\cite{he2016deep}.

     \begin{figure*}
       \centering
       \includegraphics[scale=0.51]{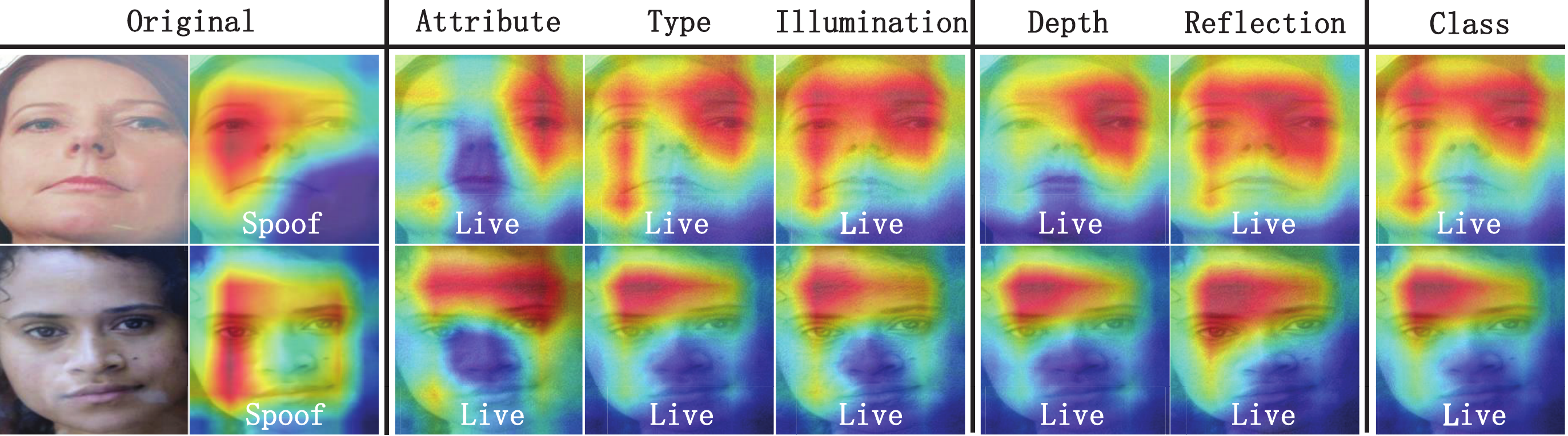}
       \caption{Visualization of adversarial attacks on different annotation vectors and geometric maps of spoofing data. For facial attributes, although the adversarial attack has perturbed the feature vector of facial attributes, it does not change the pattern of spoofing activation, different from other annotation vectors and geometric maps.}
       \label{5}
    \end{figure*}

\begin{table*}
\footnotesize
        \centering
        \caption{The accuracy of different backbone networks when their annotation vectors and geometric maps are attacked. Note that we use the decline of accuracy to represent which part is more vulnerable to adversarial attacks.}
        \begin{tabular}{cccccc}
        \Xhline{1pt}
        \multicolumn{2}{c}{Backbone Network}& VGG~\protect\cite{simonyan2014very} &
        Resnet~\protect\cite{he2016deep} & Densenet~\protect\cite{huang2017densely} & Swin Transformer~\protect\cite{liu2021swin} \\
        \Xhline{0.5pt}
        \multicolumn{2}{c}{Original Accuracy}& 0.9416 & 0.9988 & 0.9971 & 0.9989 \\
         
        \multirow{3}{*}{Annotation}& Facial Attribute & 0.7849 & 0.5799 & 0.6598 & 0.4623\\
        ~&Spoofing Type & 0.3462 & 0.2441 & 0.3838 & 0.1684\\
        ~&Illumination & 0.2736 & 0.2321 & 0.2999 & 0.2965\\
         
        \multirow{2}{*}{Geometric Map} & Depth Map & 0.4483 &0.3306 & 0.3686 & 0.3302\\
        ~ & Reflection Map & 0.6484 & 0.3343 & 0.3304 &	0.3408\\
         
        \multicolumn{2}{c}{Classification} & 0.2876 & 0.0954 & 0.2744	& 0.0962 \\
        \Xhline{1pt}
        \end{tabular}
        \label{t5}
    \end{table*}
    
    \begin{table*}
    \footnotesize
        \centering
        \caption{The changes of annotation vectors and geometric maps when the binary classification of the target model is attacked.}
        \begin{tabular}{cccccc}
        \Xhline{1pt}
        \multicolumn{2}{c}{Backbone Network}& VGG~\protect\cite{simonyan2014very} &
        Resnet~\protect\cite{he2016deep} & Densenet~\protect\cite{huang2017densely} & Swin Transformer~\protect\cite{liu2021swin} \\
        \Xhline{0.5pt}
        \multirow{3}{*}{Annotation}& Facial Attribute & 0.39 & 0.20 &0.07 & 0.20\\
        ~&Spoofing Type & 0.70 & 2.40 & 0.03 & 0.20\\
        ~&Illumination & 12.80 & 6.60 & 0.70 & 8.40\\
         
        \multirow{2}{*}{Geometric Map} & Depth Map & 4.21 & 0.10& 0.04 & 0.00\\
        ~ & Reflection Map & 2.30 & 13.60 & 1.80 & 0.40\\
        \Xhline{1pt}
        \end{tabular}
        
        \label{t6}
    \end{table*}

    \begin{table*}
    \footnotesize
        \centering
        \caption{The accuracy of different backbone networks when attacked by other networks.}
        
        \begin{tabular}{cccccccccccc}
        \Xhline{1pt}
         \multirow{2}{*}{Backbone Network}& \multicolumn{2}{c}{VGG~\protect\cite{simonyan2014very}}& &
        \multicolumn{2}{c}{Resnet~\protect\cite{he2016deep}} & & \multicolumn{2}{c}{Densenet~\protect\cite{huang2017densely}}& & \multicolumn{2}{c}{Swin Transformer~\protect\cite{liu2021swin}} \\
        \cline{2-3}\cline{5-6}\cline{8-9}\cline{11-12}
        ~&w/o SFA& with SFA & &w/o SFA& with SFA& &w/o SFA& with SFA& &w/o SFA& with SFA\\
        \Xhline{0.5pt}
        Original Accuracy& \multicolumn{2}{c}{0.9416}&  & \multicolumn{2}{c}{0.9988}&  & \multicolumn{2}{c}{0.9971} & &\multicolumn{2}{c}{0.9989} \\
        VGG~\protect\cite{simonyan2014very}& 0.6121 & \textbf{0.2876} & & 0.7863 & \textbf{0.1161} & &0.81008 &\textbf{0.3738}& & 0.2679& \textbf{0.2001}\\
        Resnet~\protect\cite{he2016deep}& 0.4212 & \textbf{0.1190}& & 0.3101& \textbf{0.0954}&  & 0.3245& \textbf{0.1394}&  & 0.4016& \textbf{0.1297}\\
        Densenet~\protect\cite{huang2017densely} & 0.4521& \textbf{0.2289}&  & 0.5025& \textbf{0.1062}&  & 0.5763&\textbf{0.2744}&  & 0.5538 & \textbf{0.2840}\\
        Swin Transformer~\protect\cite{liu2021swin} & 0.3192 & \textbf{0.1224}&  & 0.2989& \textbf{0.1018}& & 0.2296& \textbf{0.0904}&  & 0.1977& \textbf{0.0962}\\
        \Xhline{1pt}
        \end{tabular}
        \label{transfer}
    \end{table*}
    
    \noindent
    \textbf{Differences among SFA, Data Augmentation and Class Activation Map.} Since the SFA module is used to enhance the semantic information of images by introducing data prior, its function is similar to that of data augmentation and class activation map. As shown in Tab.~\ref{t2}, we select typical methods of data augmentation and class activation map (CAM)~\cite{selvaraju2017grad,chattopadhay2018grad,wang2020score,ramaswamy2020ablation,muhammad2020eigen,jiang2021layercam} to compare with SFA and take FGSM~\cite{huang2017adversarial} as the adversarial attack. To compare with the data augmentation, we not only evaluate operations separately, but also present the average of three combinations: (a) Vertical flip + Horizontal flip + Brightness, (b) Vertical flip + Horizontal flip + Rotation + Hue, and (c) Vertical flip + Horizontal flip + Rotation + Brightness + Hue. The geometric transformations made by data augmentations diversify the gradients but have little effect on the enhancement of semantic-aware features. Both CAM and SFA have studied the activation of the model towards specific classes, but CAM only considers the unique class. When SFA generates activation maps, it not only enhances the target class but also considers the opposite class.

    \noindent
    \textbf{Ablation of Texture Filter.} The CNN-based networks are biased to texture features~\cite{geirhos2018imagenet} (i.e., the discrimination results of the CNN-based models are easily effected by texture manipulation), which is the vulnerability of methods based on deep learning. To introduce this model prior into the generation process of semantic feature augmentation, we adopt the LBP as the texture filter for the SFA module. As shown in Tab.~\ref{t3}, we conduct the ablation study of the texture filter, and demonstrate that considering changing texture features can improve the success rate of different adversarial attacks.
    
\subsection{Fine-Grained Adversarial Analysis}

\noindent
\textbf{Analysis of Auxiliary Information.} Take the Resnet-18~\cite{he2016deep} trained on CelebA-spoof~\cite{zhang2020celeba} as the target model, and FGSM~\cite{huang2017adversarial} is used as the method of adversarial attacks. As shown in Tab.~\ref{t4}, SFA can improve the success attack rate of all parts except facial attributes. As shown in Fig.~\ref{5}, for facial attributes, its pattern of activation is different from other annotation vectors and geometric maps. This shows that spoofing types, illumination, depth map, and reflection map are more relevant to the ground truth decision boundary. As shown in Fig.~\ref{6}, we take the depth map as an example to interpret such differences. Considering a spoofing sample, the depth map boundary has a higher correlation with the decision boundary, so SFA can make it pass through such a decision boundary. However, the facial attribute boundary has a low correlation with the decision boundary of the live and spoofing. Even if SFA biases the example towards that decision boundary, it fails to cross such boundary of classification of the live and spoofing.

\noindent
\textbf{Analysis of Backbone Networks.} According to the latest survey~\cite{yu2022deep}, we select VGG-13~\cite{simonyan2014very}, Resnet-18~\cite{he2016deep}, Densenet-121~\cite{huang2017densely} and Swin Transformer~\cite{liu2021swin} as four representative backbone networks. The checkpoints of the target models are trained on CelebA-spoof~\cite{zhang2020celeba}, and the accuracy of each model is close to 100$\%$. Analysis of different backbone networks in the task of face anti-spoofing will be carried out in three perspectives: (a) The changes of accuracy when different annotation and geometric parts are attacked. (b) The changes of each annotation and geometric part when the binary classification of the live and spoofing is attacked. (c) We use the adversarial examples generated by one network to attack the others to evaluate the adversarial transferability of different backbone networks.
    
   \textit{\textbf{a. Adversarial Attacks on Different Annotation and Geometric Maps}} 
    
    We attack the different annotations (facial attributes, spoofing types, and illumination), geometric maps (depth and reflection), and binary classification in turn, using FGSM~\cite{huang2017adversarial} and the same max perturbation scale $\epsilon=0.2$. The changes in the accuracy of each backbone network are shown in Tab.~\ref{t5}. Through the comparison of columns, the correlation of each feature vector towards the binary classification of the live and spoofing can be reflected. The accuracy of facial attributes deserves attention while attacking facial attributes has less impact on the results of binary classification. Such a phenomenon exists in all four backbone networks. This shows that the annotation of facial attributes is redundant annotation for the binary classification of live and spoofing. Then, we analyze other attacked parts of different backbone networks from the rows. Interestingly, VGG and Densenet have better adversarial robustness when the accuracy is approximately equal. Therefore, when selecting backbone networks for specific tasks, we should handle the trade-off between accuracy and adversarial robustness.
    
    \textit{\textbf{b. Adversarial Attacks on Spoofing-Live Classification}} 
    
    As shown in Tab.~\ref{t6}, we present the changes of each annotation vector and geometric map of different backbone networks when the classification of live and spoofing is attacked by FGSM ($\epsilon=0.2$)~\cite{huang2017adversarial}. The Mean Square Error is used as a metric of change. To have better comparability among various feature vectors, the error is divided by $l_{2}$-norm of output vectors and maps. Observing the rows, the structure of VGG can enhance the correlations between different feature parts, resulting in obvious changes in other features when the binary classification is attacked. Through the comparison of columns, the parts related to light tend to change a lot, including illumination and reflection maps. According to \cite{kim2019basn} and human perception, the differences in light between live and spoofing samples are obvious, especially in spoofing types such as replay. Furthermore, for RGB images, although the depth information is used in our experiments, the change in the depth map is small. It is worth noting that in the application scenario where most existing commercial cameras collect RGB images, the way of making the networks learn to use depth information is an important problem when tackling binary classification of live and spoofing. 

    \textit{\textbf{c. The Adversarial Transferability among Different Backbone Networks}}

    We use FGSM~\cite{huang2017adversarial} and the same max perturbation scale $\epsilon=0.2$ to generate adversarial examples based on different backbone networks, and evaluate the accuracy of the other networks when they are attacked by these adversarial examples. As shown in Tab.~\ref{transfer}, the adversarial examples have some ability to transfer adversarial attacks to the other networks. With the help of our SFA module, the adversarial transferability of adversarial examples can be strengthened.
     
\section{Conclusions and Ethical Concerns}

  In this paper, we propose a novel framework to expose the fine-grained adversarial vulnerability of face anti-spoofing models. Our semantic feature augmentation (SFA) module is able to provide more discrimination-related gradients and increase the attack success rate by nearly 40$\%$ on average. We use these tools to evaluate three annotations, two geometric maps, and four backbone networks, drawing several meaningful and practical results. These novel perspectives can help our community to select annotations, geometric information, and backbone networks with better adversarial robustness when solving the task of face anti-spoofing, so as to improve the reliability of biometric authentication systems. In the future, we will explore adversarial learning based on our adversarial attacks.
    
    \noindent
   \textbf{Ethical Concerns.} Before launching defense, studying attacks is significantly necessary. This paper guides the adversarial examples as a tool to analyze face anti-spoofing models. We aim to improve the positive impact of the adversarial examples. The methods proposed in our paper can deepen our understanding of data and models, instead of undermining the security of current face recognition systems.

\section{Acknowledgments}
\vspace{-0.2cm}
This work is supported by the National Key Research and Development Program of China under Grant No. 2021YFC3320103, the National Natural Science Foundation of China (NSFC) under Grant 61972395, 62272460, a grant from Young Elite Scientists Sponsorship Program by CAST (YESS), and sponsored by CAAI-Huawei MindSpore Open Fund.

%%%%%%%%% REFERENCES
{\small
\bibliographystyle{ieee_fullname}
\bibliography{egbib}
}

\end{document}